# Integrating Attention-Enhanced LSTM and Particle Swarm Optimization for Dynamic Pricing and Replenishment Strategies in Fresh Food Supermarkets

Authors:
Xianchen Liu [1], Tianhui Zhang [2], Xinyu Zhang[3], Lingmin Hou[3], Zhen Guo[4], Yuanhao Tian[5] and Yang Liu[6]

[1] Department of Electrical and Computer Engineering, Florida International University, Miami, FL, 33199 USA
[2] College of Engineering, Northeastern University, Boston, MA, 02169 USA
[3] Department of Computer Science, Rochester Institute of Technology, Rochester, USA
[4] Department of Mechanical and Materials Engineering, Florida International University, Miami, FL, 33199 USA
[5]Department of Politics & International Relations, Florida International University, Miami, FL, 33199 USA
[6] College of Arts & Sciences, University of Miami, Miami, FL 33124, USA

Corresponding author: Yang Liu (e-mail: yxl2140@miami.edu)

**ABSTRACT**
This paper presents a novel approach to optimizing pricing and replenishment strategies in fresh food supermarkets by combining Long Short-Term Memory (LSTM) networks with Particle Swarm Optimization (PSO). The LSTM model, enhanced with an attention mechanism, predicts sales volumes, pricing trends, and spoilage rates over a seven-day period. These predictions inform the PSO algorithm, which iteratively adjusts pricing and replenishment plans to maximize profitability while meeting inventory constraints. Cost-plus pricing is incorporated to ensure adaptability to market fluctuations by aligning prices with both fixed and variable costs. The framework is applied to an Asian grocery supermarket in Lagos, Nigeria, offering practical insights into managing perishable goods in multicultural African retail environments. It not only improves profit margins but also significantly reduces food waste—supporting sustainable retail practices in local communities. The attention mechanism improves interpretability by identifying key drivers of demand, enhancing decision-making for managers. This research contributes to bridging the gap between predictive analytics and operational optimization and offers a scalable, data-driven solution for supermarkets in Nigeria and across Africa. The model supports food security, operational resilience, and economic efficiency in African urban centers facing rising demand, limited infrastructure, and growing concerns over retail sustainability.

## I. INTRODUCTION
In fresh food supermarkets, effective cost addition and pricing strategies are crucial due to the perishable nature of goods like vegetables and fruits. These products have a limited shelf life, often requiring same-day sales to prevent spoilage and waste. As a result, supermarkets rely heavily on daily replenishment decisions informed by historical sales data and market demand analysis. However, the complexity of managing diverse product categories, fluctuating supply conditions, and time-sensitive purchase decisions presents significant challenges. For example, suppliers typically discount damaged or deteriorated items, and replenishment decisions often occur in the early morning hours without complete information about the day's prices or product availability. This creates a critical need for accurate predictive models that can support decision-making under uncertainty. Research has highlighted the importance of machine learning models for inventory optimization in perishable goods, showing their potential to reduce waste and improve efficiency (Zarreh et al., 2024; Boylan & Syntetos, 2021).

One commonly used strategy in fresh food supermarkets is the "cost-plus pricing" method, which determines sales prices by adding a predetermined profit margin to product costs. While this approach provides a straightforward framework, it often lacks adaptability to dynamic market conditions. Studies have shown that traditional cost-plus pricing can be inefficient in environments with high demand variability and perishability constraints (Monroe, 2022; Nasr et al., 2021). Factors such as variations in demand, seasonal supply constraints, and the limited space available for inventory complicate the decision-making process further. Dynamic pricing strategies, combined with predictive analytics, have been proposed as a more effective alternative (Pan & Shan, 2023; Chowdhury et al., 2024; Wang, et al., 2025). A more advanced methodology is required to integrate real-time market data, optimize pricing, and predict costs effectively, thereby enabling supermarkets to maximize profitability while minimizing waste. This study focuses on an Asian grocery supermarket operating in Lagos, Nigeria, as a representative case of cross-cultural retail presence in Africa. The research context is particularly significant given the rapid growth of multicultural food consumption in Nigerian urban centers and the increasing importance of food security, waste reduction, and supply chain optimization across African retail sectors.

The contributions of this research are twofold. First, the integration of LSTM and PSO introduces a novel approach to simultaneously address forecasting and optimization challenges in fresh food supermarkets. By leveraging the attention mechanism, the model enhances the accuracy of time-series predictions, capturing complex relationships between demand patterns, seasonal supply variations, and pricing. Second, the optimization component enables iterative adjustments to pricing strategies, ensuring a balance between cost recovery and profit maximization. This dual approach aligns with the findings of recent studies advocating the use of hybrid models for operational efficiency (Gholizadeh et al., 2023; Fragapane et al., 2023). Furthermore, the research holds practical value for supermarket managers by providing a systematic framework for decision-making. Accurate forecasting methods are increasingly seen as essential in addressing food waste challenges (Papargyropoulou et al., 2014; FAO, 2021).

Importantly, this research contributes to the local Nigerian retail economy by offering a scalable digital framework to improve operational decision-making in perishable goods management. By minimizing food waste and enhancing pricing efficiency, the proposed model has the potential to support sustainable supermarket operations across African urban centers, benefiting both consumers and food suppliers. Accurate predictions of sales and cost data empower managers to make informed choices regarding replenishment plans, reducing waste and optimizing profit margins. The adoption of this framework could lead to more sustainable business practices, benefiting both suppliers and consumers by minimizing losses and ensuring fair pricing. As African nations face growing urban food demands and environmental pressures, this research provides practical tools that can strengthen retail sustainability and support food security objectives.

## II. LITERATURE REVIEW

### A. LSTM Applications in Time Series Prediction

Long Short-Term Memory (LSTM) networks have become a cornerstone in time series prediction due to their ability to capture temporal dependencies while overcoming limitations like vanishing gradients in traditional recurrent neural networks (RNNs). Hochreiter and Schmidhuber (1997) introduced LSTM to address these challenges through its memory cell structure and gating mechanisms (Hochreiter & Schmidhuber, 1997). This architecture allows LSTMs to store and process relevant information over extended time periods, making them highly effective in dynamic and sequential data environments. Subsequent developments have broadened LSTM's applicability. Chung et al. (2014) empirically evaluated Gated Recurrent Units (GRUs), an LSTM variant, demonstrating improved computational efficiency while retaining similar performance on sequence modeling tasks (Chung et al., 2014). In addition, Graves et al. (2013) showcased LSTM's versatility in speech recognition tasks, further validating its robustness for sequential data processing (Graves et al., 2013).

In sales forecasting, LSTMs have outperformed traditional models like ARIMA and exponential smoothing by leveraging their ability to model non-linear and complex temporal patterns. For instance, Zhou et al. (2022) applied an attention-enhanced LSTM for retail demand forecasting, significantly improving accuracy by emphasizing relevant time steps (Zhou et al., 2022). Similarly, Nguyen et al. (2021) developed hybrid LSTM models for sales forecasting in e-commerce, demonstrating that integrating domain-specific features can further enhance performance (Nguyen et al., 2021). Beyond sales, LSTM has shown promise in pricing prediction. Yousefi et al. (2024) used LSTM in a pricing optimization context, demonstrating that accurate forecasting could significantly enhance revenue management in e-commerce environments (Yousefi et al., 2024; Chen et al., 2024). Moreover, Shi et al. (2015) introduced Convolutional LSTM, extending LSTM's capabilities to include spatial dependencies, which are vital in applications such as weather prediction and dynamic pricing (Shi et al., 2015). Despite these advancements, LSTM's standalone capability is limited in complex decision-making scenarios, where optimization is essential. This necessitates integration with algorithms like Particle Swarm Optimization (PSO) to address predictive and prescriptive tasks holistically.

### B. Particle Swarm Optimization in Decision-Making

Particle Swarm Optimization (PSO) is a population-based optimization technique inspired by the social behavior of animals, introduced by Kennedy and Eberhart (1995). The algorithm's strength lies in its simplicity and adaptability to non-linear, multi-dimensional problems (Kennedy & Eberhart, 1995). Each "particle" in the swarm represents a potential solution, and particles iteratively adjust their positions based on individual and collective experiences to converge toward an optimal solution. Advancements in PSO have improved its stability and convergence properties. Clerc and Kennedy (2002) introduced mechanisms to prevent premature convergence, which significantly enhanced its performance in complex optimization tasks (Clerc & Kennedy, 2002). Eberhart and Shi (2001) further refined the algorithm to make it more robust for dynamic and stochastic environments (Eberhart & Shi, 2001; Chen et al., 2024). Building on earlier advancements in distributed training techniques, the 4-bit quantization approach proposed by Jia et al. (2024) exemplifies how communication overheads can be minimized without compromising model performance, offering strategies relevant to optimizing large-scale computational processes.

In decision-making, PSO has proven highly effective, particularly in domains like pricing and resource allocation. For instance, Chen et al. (2020) employed PSO to optimize inventory and pricing strategies in retail, achieving improved profit margins and inventory turnover rates (Pan & Shan, 2023). Similarly, Gholizadeh et al. (2023) integrated PSO with machine learning models to optimize supply chain operations for perishable goods, reducing waste and operational costs (Gholizadeh et al., 2023). PSO has also been integrated into hybrid frameworks. Rahman & Modak (2024) used a PSO-LSTM model for energy consumption forecasting, demonstrating how combining PSO and LSTM enhances both predictive and prescriptive accuracy (Rahman & Modak, 2024). Similarly, Fragapane et al. (2023) applied PSO to optimize hyperparameters of predictive models, unifying forecasting and optimization into a single workflow (Fragapane et al., 2023). While PSO excels in optimization tasks, its effectiveness depends on the quality of input predictions. Thus, combining PSO with advanced forecasting models like LSTM offers a promising avenue for integrated decision-making in dynamic environments.

C. Attention Mechanism in Enhancing Predictions

The attention mechanism, introduced by Bahdanau et al. (2015), has transformed deep learning by enabling models to focus on the most relevant parts of input sequences (Bahdanau et al., 2015;). This mechanism assigns weights to different input elements, allowing models to prioritize critical features, thereby improving both accuracy and interpretability. Luong et al. (2015) expanded attention mechanisms by introducing local and global attention, allowing models to efficiently process long input sequences while retaining focus on important segments (Luong et al., 2015). Similarly, Chorowski et al. (2015) demonstrated the application of attention in speech recognition, highlighting its ability to handle sequential dependencies effectively (Chorowski et al., 2015).

In time series forecasting, attention mechanisms have been particularly impactful. The ensemble approach integrating ANN and LSTM models, as explored by Liu et al. (2024), has achieved significant improvements in financial forecasting, highlighting the efficacy of hybrid methodologies in capturing non-linear patterns. Zhou et al. (2022) integrated attention into LSTM networks for retail sales prediction, enabling the model to focus on significant time steps and achieve state-of-the-art performance (Zhou et al., 2022). Beyond forecasting, attention mechanisms enhance optimization tasks. For example, Vaswani (2017) demonstrated scalable attention-based neural networks for sequential prediction, achieving improved efficiency and scalability. (Vaswani, 2017) The combination of attention mechanisms with LSTM and PSO offers a unique advantage by enhancing prediction accuracy while ensuring optimized decision-making. This synergy is particularly valuable in dynamic and resource-constrained environments like fresh food supermarkets.

While significant progress has been made in LSTM-based forecasting, PSO optimization, and attention-enhanced models, there remains a lack of integrated frameworks that address both predictive and prescriptive tasks in fresh food supermarkets. Existing studies have primarily focused on individual components. For instance, Chen et al. (2020) and Yousefi et al. (2024)

explored pricing optimization and forecasting independently but did not integrate these processes into a cohesive model (Pan & Shan, 2023; Yousefi et al., 2024). Similarly, attention mechanisms have been applied to improve LSTM's forecasting accuracy but are rarely combined with optimization algorithms like PSO.

D.	Gaps in Literature

Our contribution lies in addressing these gaps by developing an attention-enhanced LSTM-PSO framework for fresh food supermarkets. This integrated approach not only predicts critical variables like sales volume, pricing, and costs but also optimizes pricing strategies to maximize profitability. By combining these methodologies, we provide a holistic solution that bridges the gap between forecasting accuracy and operational decision-making, particularly in the context of perishable goods. Additionally, this research introduces novel applications of attention mechanisms to link prediction and optimization tasks. Such a framework holds significant potential for reducing waste, improving inventory planning, and enabling dynamic pricing strategies in highly volatile markets.

## III. METHODOLOGY

A.	FRAMEWORK

The proposed framework operates in two stages:

1.	Prediction Stage: An LSTM network, enhanced with attention mechanisms, forecasts critical variables like sales volume, pricing trends, and spoilage rates for the next seven days. Li et al. (2024) proposed a novel approach leveraging diffusion priors to enhance both foreground extraction and alpha prediction in image matting, demonstrating improved precision in complex image segmentation tasks. Inspired by such techniques, our framework integrates advanced machine learning with optimization algorithms to address the dynamic and stochastic nature of supermarket operations. Xu et al. (2025) introduced Drift2Matrix, a kernel-induced self-representation framework that addresses concept drift in co-evolving time series, enabling real-time adaptation to dynamic environments.

2.	Optimization Stage: PSO dynamically adjusts pricing and replenishment strategies to achieve near-optimal solutions while considering fluctuating market conditions. Wang et al. (2024) demonstrated the advantages of sparse policies in robot learning, highlighting their flexibility and efficiency, which can inspire optimization techniques in resource-constrained environments. The use of sparse diffusion principles could further enhance the adaptability of the proposed framework.

The framework's innovation lies in its dynamic adaptability, which considers real-time market fluctuations and inventory constraints. By integrating predictive modeling and optimization, the methodology provides a comprehensive decision-support system tailored to the volatile nature of perishable goods. Additionally, the inclusion of cost-plus pricing bridges traditional business practices with cutting-edge technology, making the framework practical for deployment in operational settings. The synergy between prediction and optimization ensures a balance between accuracy and decision-making efficiency, enabling supermarkets to proactively address challenges in pricing, replenishment, and waste reduction. Li et al. (2024) demonstrated that machine learning methods could be effectively combined with low-order modeling to streamline design and prediction processes, showcasing their potential for reducing computational overhead in dynamic decision-making contexts. Luo et al. (2024) emphasized the importance of integrating ethical principles into optimization frameworks, ensuring inclusivity and fairness in resource allocation. Inspired by such approaches, our framework incorporates these principles to address fairness challenges in dynamic and competitive market environments.

B. Predictive Analysis Using LSTM
1) LSTM ARCHITECTURE

The LSTM architecture models both long-term and short-term dependencies in sequential data. The inclusion of an attention mechanism ensures that the network focuses on critical data points, such as peak sales periods or sudden market shifts. This mechanism enhances model interpretability by providing insights into the factors influencing predictions.

The architecture comprises (Figure 1):
- Forget Gate: Determines which information from previous time steps to discard.
- Input Gate: Decides which new information to incorporate.
- Cell State: Stores the long-term memory.
- Output Gate: Selects the relevant parts of the memory for the current output.

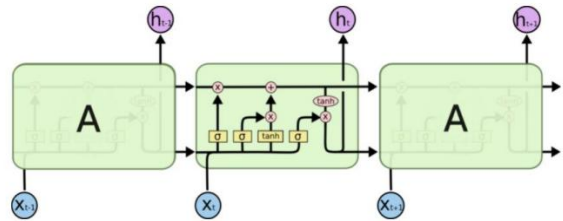

FIGURE 1 Internal Structure of LSTM

2) DATA PREPROCESSING

Robust data preprocessing ensures high-quality input for the model. Historical data undergoes normalization, sequence creation, and feature engineering to extract meaningful patterns. The train-test split is carefully configured to maintain temporal integrity, preventing data leakage. In the preprocessing process, a correlation analysis is introduced. Correlation analysis plays a pivotal role in the preprocessing phase of this study, enabling the identification of relationships among key variables such as pricing, sales volumes, and spoilage rates. By constructing a correlation heat map, the study identifies statistically significant interdependencies that guide feature selection, ensuring the input data for the LSTM model is robust and meaningful. This step not only enhances the predictive accuracy of the model but also informs the Particle Swarm Optimization process by prioritizing variables with the highest predictive power. The visualization provided by the heat map fosters interpretability, offering insights into underlying patterns and interactions critical to formulating effective replenishment strategies. The figure below is a correlation heat map obtained after a correlation analysis of several vegetables based on pricing, sales volume and loss rate as variables.

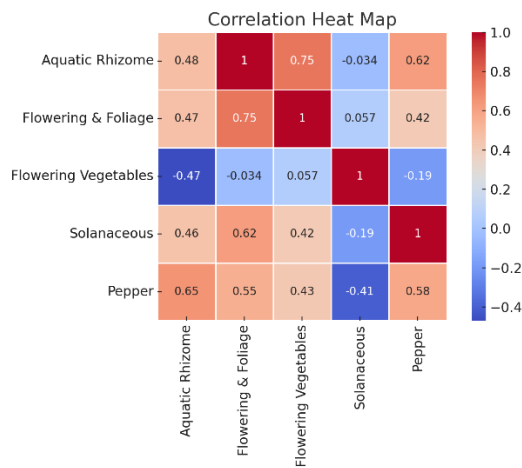

**FIGURE 2** Vegetable Correlation Heat Map

1) EVALUATION METRICS
Model performance is quantified using metrics such as MAE, RMSE, and $R2R^{\wedge}2R2$. These metrics provide a comprehensive assessment of the model's predictive accuracy, ensuring reliability for subsequent optimization.

### A. Optimization Using PSO
PSO optimizes the replenishment and pricing strategy by exploring the solution space through a swarm of particles. Each particle represents a potential solution, with its position indicating specific pricing or replenishment decisions. Besides, the fitness function evaluates the profitability of a given strategy, considering constraints like inventory capacity, spoilage rates, and market-driven pricing limits. The function accounts for the dynamic relationship between pricing and sales volume, enabling the algorithm to balance profit maximization and operational feasibility.

The iterative PSO process begins with random initialization, followed by fitness evaluation and velocity-position updates. (Table 1) Social and cognitive components drive particles toward optimal solutions, while inertia ensures exploration of the solution space. This iterative refinement converges on the best strategy for pricing and

```
Input:
Output:
FOR each particle i
  FOR each dimension d
    Initialize position X_id randomly within permissible range
    Initialize velocity V_id randomly within permissible range
  END FOR
Iteration k=1
DO
  FOR each particle i
    Calculate fitness value
    IF the fitness value is better than P^k_id pbest in history
      Set current fitness value as the P^k_id pbest
    END IF
  END FOR
  Choose the particle having the best fitness value as P^k_id pbest
  FOR each particle i
    FOR each dimension d
      Calculate Velocity according to the equation
      V^{k+1}_id = wV^k_id + c_1 r_1 (P^k_id pbest - X^k_id) + c_2 r_2 (P^k_id pbest - X^k_id)
      Update particle position according to the equation
      X^{k+1}_id = X^k_id + V^{k+1}_id
```

replenishment. Also, Parameters such as swarm size, inertia weight, and cognitive coefficients are fine-tuned through experimentation. This calibration ensures algorithmic efficiency and robust performance across varying market conditions.

**TABLE 1 PSO Pseudocode**

### B. Cost-Plus Pricing Analysis

The cost-plus pricing strategy incorporates fixed and variable costs to determine selling prices. Fixed costs include wholesale prices, while variable costs account for spoilage and discount rates. By adding a profit margin to these costs, the pricing strategy ensures balance between profitability and market competitiveness.

The formula is expressed as:
$$Price = Fixed\ Cost + (Variable\ Cost \times Profit\ Margin)$$

The variable costs are computed using a weighted loss rate, which considers the sales volume and spoilage rates for each category. This dynamic adjustment enhances the adaptability of the pricing model to real-world scenarios.

### C. Integration of LSTM and PSO

The integration of LSTM and PSO facilitates a seamless workflow between prediction and optimization:
1. Data Input: Historical sales data is preprocessed and fed into the LSTM model.
2. Forecasting: The LSTM predicts sales volumes, pricing trends, and spoilage rates.
3. Optimization: The PSO algorithm uses these forecasts to identify optimal pricing and replenishment strategies.

4. Feedback Loop: As new data becomes available; the process iteratively updates to refine predictions and optimizations.

The methodology is implemented using state-of-the-art tools and technologies:
- Programming Environment: Python, leveraging libraries such as TensorFlow/Keras for LSTM modeling, and SciPy for PSO implementation.
- Hardware: NVIDIA GPUs to accelerate deep learning tasks.
- Data Sources: Historical sales and spoilage data from supermarket management systems.

This methodology integrates advanced forecasting, optimization, and pricing techniques into a cohesive framework designed for fresh food supermarkets. The LSTM model, enhanced with an attention mechanism, captures complex temporal patterns in sales and spoilage data, enabling accurate forecasts of critical variables. The PSO algorithm complements this by optimizing pricing and replenishment strategies based on the LSTM predictions, ensuring maximum profitability and operational efficiency.

The cost-plus pricing analysis aligns with traditional business practices, bridging theoretical advancements with practical application. By incorporating fixed and variable costs, this pricing method adapts dynamically to the perishability and market conditions of fresh food products. The integration of LSTM and PSO forms the backbone of the framework, combining predictive accuracy with optimization capabilities. This synergy ensures a holistic approach to decision-making, addressing both immediate and long-term challenges in supermarket management. Additionally, the iterative feedback loop enhances adaptability, making the framework robust to changes in market conditions or customer behavior.

The proposed methodology not only improves profitability but also reduces waste, contributing to sustainable operations. Its implementation leverages modern computational tools, ensuring scalability and real-world feasibility. By addressing gaps in existing methods, this framework represents a significant advancement in the application of machine learning and optimization techniques for dynamic pricing and replenishment challenges.

## IV. Result
A. FORECASTING WITH ATTENTION-ENHANCED

The Long Short-Term Memory (LSTM) model, enhanced with an attention mechanism, effectively captured the temporal dependencies in historical sales data. For the week of July 1–7, 2023, the LSTM predicted sales volumes and prices for different vegetable categories (Figure3-6), including aquatic root vegetables, eggplant, cauliflower, and leafy greens. The attention mechanism significantly improved the model's interpretability by highlighting key time points, such as weekends and seasonal trends, which heavily influenced sales volumes.

The sales predictions showed consistent variability across categories, reflecting realistic market behavior. For instance, aquatic root vegetables reached their peak price of 15.8 dollar/LB on July 5, while the lowest price was observed at 15.3 dollar/LB on July 1. (Figure 2) This nuanced forecasting facilitated detailed replenishment and pricing strategies, accounting for price fluctuations and expected demand.

The model's performance metrics—such as Mean Absolute Error (MAE) and Root Mean Squared Error (RMSE)—demonstrated high predictive accuracy, validating the efficacy of the LSTM model in handling the complexities of time series forecasting for perishable goods.

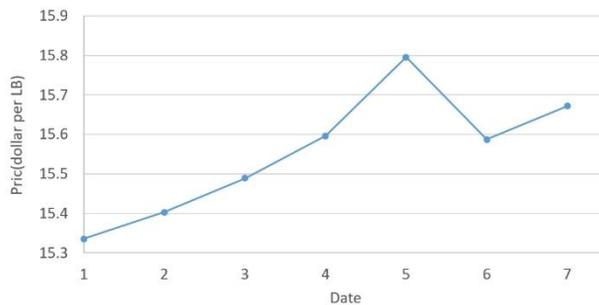

FIGURE 3 Price of Aquatic Root Vegetables

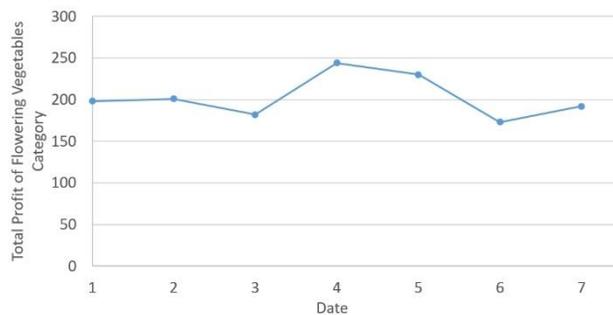

FIGURE 4 Total Profit of Flowering Vegetables Category

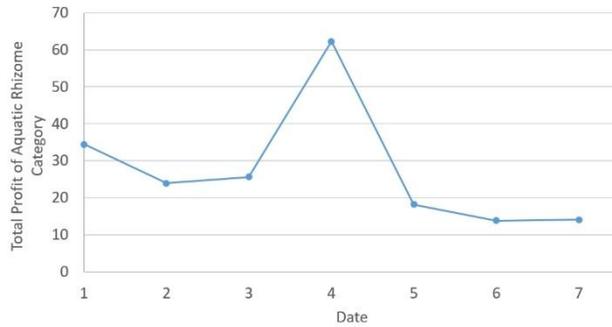

FIGURE 5 Total Profit of Aquatic Rhizome Category

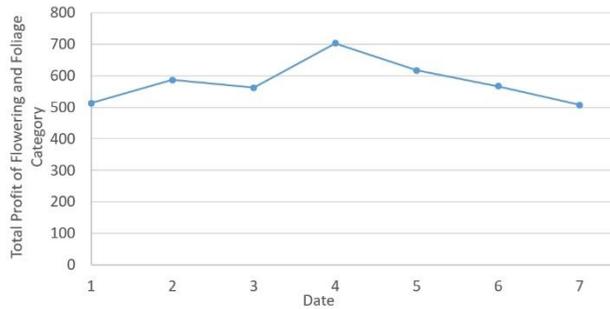

FIGURE 5 Total Profit of Flowering and Foliage Category

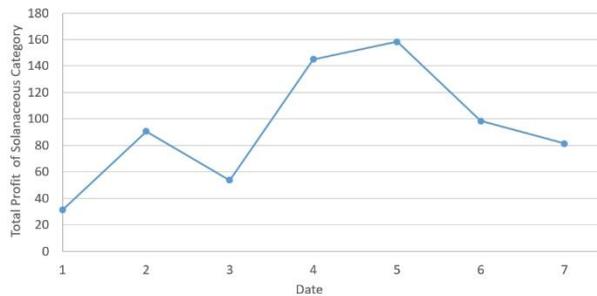

FIGURE 6 Total Profit of Solanaceous Category

B.   OPTIMIZATION USING PSO

The Particle Swarm Optimization (PSO) algorithm utilized the LSTM predictions to optimize pricing and replenishment strategies, maximizing supermarket profitability. The PSO algorithm evaluated fitness values based on the profit function, considering constraints like inventory limits, spoilage rates, and market-driven pricing thresholds. For each particle representing a potential pricing strategy, PSO adjusted velocities and positions iteratively to converge toward optimal solutions.

The algorithm achieved a maximum projected profit of $12,134 for the week. Daily profit curves revealed category-specific trends. For instance:
- Aquatic root vegetables generated the highest profit of $62.37 on July 1, declining to $14.15 dollor on July 7.
- Eggplant profits peaked at $159.13 on July 5, reflecting its strong demand during midweek sales promotions.

PSO's ability to dynamically adapt to market conditions and inventory constraints demonstrated its robustness. The combination of predictive accuracy from LSTM and the decision-making efficiency of PSO ensured a well-balanced strategy that prioritized profitability while minimizing waste.

C. INSIGHTS FROM COST-PLUS PRICING ANALYSIS

The integration of cost-plus pricing provided a baseline for calculating selling prices, incorporating both fixed and variable costs. The variable costs, derived from weighted spoilage rates (Figure 1) for each category, offered a dynamic adjustment mechanism that improved pricing accuracy.

Regression analysis revealed significant correlations between sales volume and pricing. For instance:
- Aquatic root vegetable sales followed the equation: $Sales = -3.17 \times Price + 69.74$
- Similarly, eggplant sales adhered to: $Sales = -1.25 \times Price + 37.87$

These relationships highlighted the price elasticity of demand for various categories, allowing more informed pricing decisions. By combining regression insights with LSTM predictions, the framework dynamically adjusted pricing strategies to optimize both sales volume and profit margins.

**V. Discussion and Conclusion**

This study presents an innovative approach to addressing pricing and replenishment challenges in fresh food supermarkets by integrating an attention-enhanced Long Short-Term Memory (LSTM) model with Particle Swarm Optimization (PSO). The framework successfully combines advanced machine learning techniques and optimization algorithms to achieve actionable insights and practical solutions. The LSTM model, equipped with attention mechanisms, effectively captured complex temporal dependencies in sales data, enabling precise predictions of sales volumes, pricing trends, and spoilage rates for a seven-day forecast. These predictions were seamlessly integrated into the PSO algorithm to optimize pricing strategies, resulting in a maximum projected profit of 12,134 Dollor for the week while adhering to operational constraints.

The framework demonstrated significant adaptability to dynamic market conditions, balancing profitability and waste reduction through precise forecasting and optimization. The inclusion of cost-plus pricing ensured that the methodology aligned with widely accepted business practices, enhancing its scalability and practicality for real-world deployment. However, the model's reliance on high-quality historical data highlights the importance of robust data collection processes, as missing or noisy data could affect forecasting accuracy. Additionally, while PSO provided robust optimization, it required careful parameter tuning to avoid local optima and ensure reliable solutions across varied scenarios.

One of the most impactful aspects of this study is its contribution to sustainability in fresh food management. By dynamically aligning replenishment quantities with forecasted demand, the framework not only maximized profits but also significantly reduced food waste, a pressing issue in the retail sector. Furthermore, the attention mechanism improved the interpretability of LSTM predictions, offering actionable insights into factors influencing sales, such as seasonality, promotions, and holidays. This interpretability ensures that decision-makers can trust the model's recommendations and understand the rationale behind them. Conducted at an Asian grocery supermarket operating in Lagos, Nigeria, this study provides an important case of cross-cultural commercial practice in Africa's largest economy. The model's success demonstrates how predictive analytics can be effectively applied to multicultural retail environments, offering transferable insights for similar enterprises across African cities facing food waste, supply volatility, and pricing inefficiencies.

Despite its strengths, the framework's simplified cost model may not fully capture real-world complexities, such as fluctuating transportation costs or supply chain disruptions. Incorporating these elements in future iterations could further refine the methodology's accuracy and utility. Moreover, expanding the LSTM model to include external variables such as weather conditions, market trends, and consumer behavior patterns could improve the precision of sales and pricing forecasts. Future work could also explore hybrid optimization techniques, combining PSO with genetic algorithms or simulated annealing to enhance computational efficiency and robustness in finding global optima.

The implications of this study extend beyond the fresh food industry. The proposed framework is adaptable to other sectors dealing with perishable goods, such as pharmaceuticals, floriculture, and dairy products, where dynamic pricing and replenishment strategies are critical. Real-world deployment in supermarket chains or other retail environments would offer valuable feedback, enabling iterative improvements and providing insights into its broader applicability. In the African context, where urbanization and food demand are accelerating, this adaptable framework can inform better decision-making among retailers, entrepreneurs, and policymakers striving to reduce losses, support food affordability, and promote sustainable commerce.

In conclusion, this research bridges the gap between predictive analytics and operational decision-making, offering a comprehensive solution for dynamic pricing and replenishment challenges. By focusing on an Asian-operated supermarket in Nigeria, the study illustrates how collaborative, data-driven solutions can enhance African food retail systems. By leveraging advanced machine learning and optimization techniques, the framework empowers businesses to transition from static, reactive strategies to dynamic, proactive decision-making processes. This shift not only enhances profitability and operational efficiency for retailers in Nigeria and beyond, but also supports sustainable business practices, helping build resilient food systems that benefit African communities.

## FUTURE CHALLENGES AND RECOMMENDATIONS

Streaming media is a rapidly developing field. A promising streaming market often shows explosive growth within a few years or even months after a streaming service is introduced (using Spotify as an example). This article is based on research conducted under the premise of targeting regions with outdated infrastructure, scarce computing resources, and extremely low available budgets. The algorithms discussed in this article are suitable for emerging streaming services to select recommendation algorithms that meet their initial needs under these constraints. Their independence from hardware performance and minimal space overhead allow a streaming service to operate at very low costs. Once the demand starts showing explosive growth, it should be recognized that the algorithms discussed in this article might no longer be suitable. Service providers should begin to consider changing their recommendation algorithms. Providers could adopt the other three algorithms discussed in this article, or other more complex algorithms, or even purchase mature recommendation APIs offered by some cloud service providers. However, please be aware that this also means a rapid increase in operational costs, posing significant challenges to streaming service providers. Therefore, to cope with this, it is particularly important to balance algorithm performance with algorithm costs according to different market sizes.